\documentclass[12pt,psfig]{article}
\usepackage{graphicx}
\usepackage{float}
\usepackage{subcaption}
\usepackage{amsmath,amssymb,mathtools,bbm}
\usepackage{bbold}
\usepackage{color}
\usepackage{tikz}
\usetikzlibrary{shapes}
\usepackage{booktabs}
\usepackage{multirow}
\usepackage{longtable}
\usepackage{pgfplots}
\pgfplotsset{compat=1.17}
\usepackage{subeqnarray}
\usepackage{dsfont}
\usepackage{changepage}
\usepackage{hyperref}
\usepackage{cleveref}
\crefname{equation}{}{} % The convention is to use () and no name for equations
\crefname{figure}{fig.}{figs.}

\tikzset{cross/.style={cross out, draw=black, minimum size=2*(#1-\pgflinewidth), inner sep=0pt, outer sep=0pt},
cross/.default={1pt}}

\usepackage[symbol]{footmisc} %footnote symbol

\usepackage[bibstyle= nature, citestyle=numeric-comp, sorting=none]{biblatex}

\begin{document} 

\begin{center}

\begin{Large} 
\noindent Variational encoder geostatistical analysis (VEGAS) with an application to large scale riverine bathymetry \\
\end{Large}

\vspace{0.7cm}
\begin{small}
\noindent Mojtaba Forghani$^1$\footnote[1]{email: mojtaba@stanford.edu}, Yizhou Qian$^2$, Jonghyun Lee$^3$, Matthew Farthing$^4$, Tyler Hesser$^4$, \newline Peter K\@. Kitanidis$^{2,5}$, and Eric F\@. Darve$^{1, 2}$ \\
\end{small}

\vspace{0.3cm}
\begin{footnotesize}
$^1$ Department of Mechanical Engineering, Stanford University, CA \\
$^2$ Institute for Computational and Mathematical Engineering, Stanford University, CA\\
$^3$ Department of Civil and Environmental Engineering and Water Resources Research Center, University of Hawai`i at M\=anoa, Honolulu, HI \\
$^4$ U.S\@. Army Engineer Research and Development Center, Vicksburg, MS\\
$^5$ Department of Civil and Environmental Engineering, Stanford University, CA
\end{footnotesize}

\end{center}

\begin{abstract}
Estimation of riverbed profiles, also known as bathymetry, plays a vital role in many applications, such as safe and efficient inland navigation, prediction of bank erosion, land subsidence, and flood risk management. The high cost and complex logistics of direct bathymetry surveys, \textit{i.e.}, depth imaging, have encouraged the use of indirect measurements such as surface flow velocities. However, estimating high-resolution bathymetry from indirect measurements is an inverse problem that can be computationally challenging. Here, we propose a reduced-order model (ROM) based approach that utilizes a variational autoencoder (VAE), a type of deep neural network  with a narrow layer in the middle, to compress bathymetry and flow velocity information and accelerate bathymetry inverse problems from flow velocity measurements. In our application, the shallow-water equations (SWE) with appropriate boundary conditions (BCs), \textit{e.g.}, the discharge and/or the free surface elevation, constitute the forward problem, to predict flow velocity. Then, ROMs of the SWEs are constructed on a nonlinear manifold of low dimensionality through a variational encoder. Estimation with uncertainty quantification (UQ) is performed on the low-dimensional latent space in a Bayesian setting. We have tested our inversion approach on a one-mile reach of the Savannah River, GA, USA. Once the neural network is trained (offline stage), the proposed technique can perform the inversion operation orders of magnitude faster than traditional inversion methods that are commonly based on linear projections, such as principal component analysis (PCA), or the principal component geostatistical approach (PCGA). Furthermore, tests show that the algorithm can estimate the bathymetry with good accuracy even with sparse flow velocity measurements.
\end{abstract}

{\it keywords}: deep learning, inverse modeling, reduced-order models, riverine bathymetry, variational encoder, Bayesian estimation 

\section{Introduction}
\label{intro}

High-resolution riverine bathymetry plays an essential role in many practical applications such as the study of riverine morphodynamics, safe and efficient maritime transportation, aquatic habitat management, and flood risk management~\cite{overdeep, morph, Casas, Westaway, Lane}. However, direct high-resolution bathymetric surveys, commonly performed via wading or watercraft-mounted multi-beam sonar equipment~\cite{Casas, Marcus_mapping}, are time consuming and costly for long river reaches~\cite{PCA_DNN_Hojat}. Therefore, several remote sensing techniques have been used in the literature to obtain informative flow properties from which the bathymetry can be estimated~\cite{Emery, Inference}. These include airborne bathymetric LiDAR systems~\cite{Marcus_mapping, Hilldale, McKean, DepthLearn}, multi-spectral imagery~\cite{Misra, Wang}, satellite-derived bathymetry (SDB) using Google Earth engine~\cite{Google}, measurement of water surface elevation~\cite{Inference, Yoon}, measurement of surface velocity through GPS drifters~\cite{Emery} or particle image velocimetry with digital video cameras~\cite{Muste}, and thermal imagery~\cite{Puleo, Lima}. In this work, we estimate the bathymetry using surface flow velocity measurements, since they are relatively easy to acquire at a low cost in all river conditions (\textit{e.g.}, not sensitive to high turbidity, high or low flows). Surface velocities are sensitive to river depth (see~\cite{Lee_Hojat, neyshabur2017exploring}), which can be captured through empirical relationships or standard hydrodynamic models like the two-dimensional shallow water equations (SWEs; \cite{Landon, Wilson})---the partial differential equations (PDEs) that calculate depth and depth-averaged flow velocities from bathymetries and the boundary conditions (BCs).

Our focus in this paper is on real-time bathymetry estimation using a SWE model, since it can produce high-resolution estimation of the river depth using riverine hydrodynamics instead of the mean cross-sectional depth estimates obtained from empirical relationships~\cite{Lee_Hojat}. We will assume that the underlying river bed topography is invariant and the river flow is assumed to be at quasi-steady state during the flow velocity measurement campaign (collected in a short time, \textit{e.g.}, over a day), which are typical conditions in bathymetry estimation problems~\cite{Landon, Wilson, Lee_Hojat}. These flow velocity observations can then be inverted through the SWEs model in order to estimate the bathymetry, which we treat as an ``inverse problem.''

In principle, the riverine bathymetry identification problem is underdetermined, meaning that usually there exist multiple possible configurations of bathymetry that are consistent with the observations (\textit{i.e.}, flow velocity measurements)---the ill-posedness of inverse problems~\cite{Lee_Hojat}. Beyond that, the inversion process must also account for the uncertainty introduced not only by the measurement error but also by the imperfect representation of riverine hydrodynamics in the numerical SWEs model. In addition to information contained in the data, inversion techniques typically utilize prior knowledge that reflects understanding of the bathymetry in order to weigh possible solutions among those consistent with data. These solutions can be evaluated in a probabilistic way, which is commonly treated in a Bayesian framework~\cite{Bishop} by finding the probability density of feasible solutions that satisfy both model fitting and prior information~\cite{Kaipio}. For example, prior information is described through the statistical model in which unknown properties are distributed randomly with prescribed mean and spatial covariance functions, which form a probability density function (PDF).

As the volume of available data~\cite{Gleason} for high-resolution bathymetry imaging becomes larger, high-resolution SWEs computational models should be implemented in order to capture small-scale spatial variability of the river bottom topography. Two main computational bottlenecks arise when gradient-based optimization approaches like variational data assimilation~\cite{Strub} are applied to high-dimensional and/or data-intensive problems: the high cost of construction of the Jacobian (sensitivity) matrix and the computational and storage costs for the dense prior covariance matrix. In particular, the computational costs associated with the Jacobian matrix construction amount to a number of simulation model runs proportional to the number of observations (with the adjoint-state method~\cite{Cacuci}) or the number of unknowns~\cite{SunSun}, which would be the most expensive part in the bathymetry estimation problem considered here.

To address these issues, ensemble-based methods~\cite{Landon, wilson2010data} have been widely used for the riverine bathymetry estimation problem because of their Jacobian-free implementation. For example, \cite{Wilson} presents an ensemble-based bathymetry estimation approach with the Regional Ocean Modeling System (ROMS) in which the river depth is estimated using synthetic flow velocity observations. Using the same approach, \cite{Landon} assimilates velocity measurements obtained from drifters, in order to estimate riverine bathymetry. However, the success of the ensemble-based method for the riverine bathymetry problem strongly depends on the choice of the initial ``guess,'' \textit{i.e.}, prior mean and covariance. Furthermore, the ensemble-based method typically suffers from spurious model correlation and ensemble spread reduction that require the so-called covariance localization~\cite{Hamill, Lee_Hojat}, especially when insufficient ensemble sizes are being used~\cite{Lee_Hojat}. It has been shown that the ensemble-based method requires ensembles with many realizations when assimilating a larger number of observations~\cite{kokk, Lee_Hojat}.

The principal component geostatistical approach (PCGA)~\cite{PCGA_Kitanidis, PCGA_Lee, Lee_Hojat} is another technique that, similar to ensemble-based methods, employs low-rank approximations of the prior covariance to avoid direct Jacobian matrix computation. However, unlike ensemble-based methods that use samples from the prior distribution, PCGA uses the eigendecomposition of the covariance matrix, and thus does not require any postprocessing (\textit{i.e.}, localization). Additionally, PCGA implements successive iterations to achieve a better linearization unlike standard ensemble-based methods that perform one-step analysis leading to a solution dependent on the prior guess of the bathymetry. However, both PCGA and ensemble-based methods inherently require multiple forward model executions, which can be computationally too time-consuming to perform in real time with expensive high-resolution numerical solvers~\cite{hojat2020}. Furthermore, due to the high-dimensionality of the bathymetry problem, these approaches usually require some assumptions about the spatial structure of the solution~\cite{Lee_Hojat} such as a spatial correlation modelled by multivariate Gaussian kernels, which limits the capability and generalizability of the inversion problem.

A number of recent works have used deep learning (DL)-based approaches in order to address some of the inverse modeling issues mentioned above. That is, they are capable of fast predictions by bypassing expensive numerical solvers, and they are more accurate and admit more general and complex forms of the solution (\textit{e.g.}, universal approximation theory~\cite{Bishop}). This has led to significant interest in leveraging the strength of DL for hydrology applications~\cite{Poggio, review_Sit}, such as flood prediction~\cite{runoff}, water level estimation~\cite{Jeong}, land cover classification~\cite{Abdi}, water quality monitoring~\cite{water_quality}, and water resources management~\cite{Karimi}. There has also been a significant interest in the inverse modeling community in leveraging the strength of DL-based algorithms for bathymetry estimation~\cite{Poggio}. For example, \cite{PCA_DNN_Hojat} proposed the PCA-DNN algorithm that works by combining a deep neural network (DNN) with a reduced order model (ROM) based on the principal component analysis (PCA). PCA-DNN takes flow velocities with given BCs as inputs and outputs bathymetries. In \cite{POD_SWE}, the training dataset was generated using synthetic bathymetry profiles and the corresponding flow velocities were computed via a numerical solver of the SWEs (the forward problem). An important issue in deterministic approaches such as PCA-DNN is that using a deterministic map from velocity to bathymetry instead of solving the inverse modeling problem with a statistical framework limits the capability of these techniques to a specific set of measurements. Furthermore, the ill-posedness of the inverse problem leads to higher sensitivity of the inverse map to measurement noise.

In this paper, we propose a technique, referred to as variational encoder geostatistical analysis (VEGAS), \textbf{that incorporates both a statistical (Bayesian) approach to the inversion and a DNN-based ROM.} Our Bayesian approach attempts to find the solution of the inverse problem through an optimization process, whose iterations are performed via a computationally cheap ROM surrogate of the forward problem~\cite{ROM_optim, ROM_Willcox}. In particular, we use a supervised variational encoder (SVE)~\cite{SWE_forward} as a ROM of the SWEs. Then, we find the maximum a posteriori (MAP) estimate of the bathymetry (the optimization process) via the optimal reduced-dimension variables (latent space) of the trained SVE (see Section \ref{meth} for more details). Performing the Bayesian steps in the low-dimensional space (latent space) of the ROM instead of the original high-dimensional space of the bathymetry/velocity reduces the computational cost of the inversion significantly. Such variational encoder-type structures have been used previously as a part of a Bayesian framework to infer the prior distribution, for example, when modeling unsteady fluid flows~\cite{Akkari}. Since in our approach the ROM is combined with a Bayesian framework, it is capable of providing a posteriori PDF of the bathymetry. This is unlike deterministic methods such as PCA-DNN~\cite{PCA_DNN_Hojat}, whose prediction is deterministic and based on an inverse map. Furthermore, since our Bayesian approach models the observation uncertainty explicitly as part of its framework, it is robust to the measurement noise. Our Bayesian view to the inversion problem is similar to that of PCGA~\cite{PCGA_Lee, PCGA_Kitanidis}. However, PCGA uses a linear dimension reduction technique (similar to \cite{PCA_DNN_Hojat} and other linear ROM-based methods), while our ROM is based on a data-driven nonlinear dimension reduction, which has the potential to be more efficient and generalizable. Furthermore, PCGA uses numerical solvers to solve the forward problem (SWE) during its iterations, while here we use a DNN-based fast surrogate solver, which is orders of magnitude faster than traditional numerical solvers. This makes the inversion process possible without access to fast graphics processing units (GPUs) and special purpose solvers~\cite{shallow_GPU}. VEGAS can make online predictions possible both due to the computational advantages of the DNNs and the use of a ROM, which acts as a low-dimensional surrogate of the numerical solver of the forward model. % Note that we also perform the Bayesian steps in the low-dimensional space (latent space) of the ROM instead of the original high-dimensional space of the bathymetry/velocity, which additionally reduces the computational cost of the inversion.

The remainder of the paper is as follows. In Section \ref{meth}, we provide a brief overview of the methodology used in our work, that is, the SVE (forward solver) and the Bayesian inversion process (inverse solver). In Section \ref{data_prep}, we discuss the process being used to generate the data, such as bathymetries, BCs, and flow velocities that will be provided to the DNN. In Section \ref{result_global}, we provide the results of applying our inversion algorithm on different datasets and discuss the results. Finally, in Section \ref{conclusion} we discuss the major findings from the work and consider potential future directions.

\section{Methodology}
\label{meth}

In this work, first we generate a set of synthetic data that consists of data points in the form of BCs and bathymetries as inputs and flow velocities as outputs (labeled data). Specifically, as input BCs we provide the discharge at the upstream boundary and the free-surface elevation at the downstream boundary. We then use the two-dimensional (2D) shallow-water module of the U.S. Army Corps of Engineers' adaptive hydraulics (AdH) model~\cite{AdH} to solve the SWEs numerically and thus generate high-fidelity training set data for the offline stage. AdH provides a stabilized finite element (FE) approximation of depths and flow velocities on unstructured 2D meshes and uses as inputs bathymetry and BCs. Note that, in agreement with our assumptions about the data collection, the lateral geometry and bathymetry of the river have also been assumed to be fixed during the simulation~\cite{Landon, Wilson}. Once the dataset is generated, we train the SVE~\cite{SWE_forward, aaai_fast} in an \textit{offline} stage (see Section \ref{SVE_method}). The \textit{online} stage then identifies the optimal latent space variables of the SVE and provides the MAP estimate and uncertainty of the bathymetry, given the flow velocity measurements and the BCs (see Section \ref{bayes_meth}). In this work, we evaluate the performance of the VEGAS both in cases that the measurement locations are dense and sparse.

In Section \ref{bayes_meth}, we discuss the Bayesian approach (inverse problem solver; \textit{online} stage) used in our work and its relation to the DNN-based ROM (SVE), and then in Section \ref{SVE_method} we present the SVE architecture briefly.

\subsection{Inversion algorithm}
\label{bayes_meth}

We have used a Bayesian approach in this work as the inverse problem solver, in order to reconstruct the bathymetry from flow velocity observations. The forward problem can be defined in the form of the relationship
\begin{equation}
    {\bf y}= {\bf h}({\bf s}) + \boldsymbol{\varepsilon},
\label{forw}
\end{equation}
where ${\bf s}$ is the input (\textit{e.g.}, bathymetry), ${\bf y}$ is the observation (\textit{e.g.}, flow velocities), $\boldsymbol{\varepsilon}$ is the observation and model uncertainty noise such as a Gaussian distribution with mean zero $\boldsymbol{\varepsilon} \sim \mathcal{N}(\mathbf{0},{\bf R})$, where $\mathbf{R}$ is the model/observation error matrix, and ${\bf h}$ is the forward map. Here, we have assumed a noise level of $r_i= 0.05$ m/s for the velocity observation, where $r_i^2$ are diagonal terms of the matrix $\mathbf{R}$. The inverse problem equivalent of \Cref{forw} can be defined as a problem with unknown $m$-dimensional variable ${\bf s}$ (bathymetry) and $n$ (noisy) observation ${\bf y}$ (flow velocities in the two directions). The Bayes' rule allows us to evaluate a posterior distribution of ${\bf s}$ via 
\begin{equation}
p(\mathbf{s}|\mathbf{y}) \propto p(\mathbf{y}|\mathbf{s})p'(\mathbf{s})
= \int_{\boldsymbol \theta} p(\mathbf{y}|\mathbf{s})p'(\mathbf{s}|\boldsymbol\theta)p'(\boldsymbol\theta) d{\boldsymbol\theta},
\label{bayes_rule}
\end{equation}
where $p'\left(\cdot\right)$ represents the prior probability and $\boldsymbol\theta$ is a set of hyperparameters that models ${\bf s}$ in a hierarchical Bayesian framework.

Assume that we can parameterize ${\bf s}=D({\bf z},\boldsymbol\theta)$, where ${\bf z}$ is a $k$ ($\ll m$)-dimensional random variable. The variable ${\bf z}$ in the context of the inversion problem represents the low-dimensional representation of ${\bf s}$, which is equivalent to the latent space variable of the ROM (see Section \ref{SVE_method} for more details). In particular, we can constrain variable ${\bf z}$ to follow a Gaussian distribution to ensure the regularity of the latent space:
\begin{equation}
\mathbf{z} \sim \mathcal{N}(\mathbf{0},\boldsymbol\Sigma) .
\label{z_def}
\end{equation}
We also assume that $D$ is a deterministic map from ${\bf z}$ to ${\bf s}$ and $p'(\boldsymbol\theta)$ follows a delta distribution:
\begin{equation}
\boldsymbol\theta = \boldsymbol\delta(\boldsymbol\theta - \hat{\boldsymbol\theta}) .
\label{delta_dist}
\end{equation}
This allows us to rewrite \Cref{bayes_rule} in the form
\begin{align}
p({\bf z}|{\bf y}) &\propto  p({\bf y}|{\bf z})p'({\bf z}) \nonumber \\
&\propto \exp{\left(-({\bf y}- {\bf h}(G( {\bf z} )))^{\top} {\bf R}^{-1} ({\bf y} - {\bf h}(G( {\bf z}))) \right)} \exp{\left(-{\bf z}^{\top} \boldsymbol\Sigma^{-1} {\bf z} \right)},
\label{bayes_rule_2}
\end{align}
where ${\bf s}=G(\mathbf{z})$ is the map from the low-dimensional latent space ${\bf z}$ to the unknown variable space (\textit{e.g.}, bathymetry). Finally, the MAP estimate of \Cref{bayes_rule_2} becomes
\begin{equation}
\mathbf{z}_\text{map} =  \arg\min_{\mathbf{z}}{\left((\mathbf{y}-{\bf h}(G(\mathbf{z})))^{\top}\mathbf{R}^{-1}(\mathbf{y}-{\bf h}(G(\mathbf{z}))) + \mathbf{z}^{\top}\boldsymbol\Sigma^{-1}\mathbf{z}\right)}.
\label{z_map}
\end{equation}
The MAP estimate can be determined by the Gauss-Newton method with iterative linearizations~\cite{PCGA_Kitanidis}. For iteration count $l$ and step size $\alpha$, we have
\begin{align}
    \mathbf{z}^{l+1} &= \mathbf{z}^{l} + \alpha \left( \boldsymbol\Sigma^{-1} + (\mathbf{J}^l)^{\top}\mathbf{R}^{-1}\mathbf{J}^l \right)^{-1}\left(\boldsymbol\Sigma^{-1} {\bf z}^{l} + (\mathbf{J}^l)^{\top}\mathbf{R}^{-1}\left(\mathbf{y} - {\bf h}(G(\mathbf{z}^{l})) \right) \right) \nonumber\\
     &= (1-\alpha) \mathbf{z}^{l} + \alpha \boldsymbol\Sigma \mathbf{J}^l \left( \mathbf{J}^l \boldsymbol\Sigma (\mathbf{J}^l)^{\top} + \mathbf{R} \right)^{-1}(\mathbf{y}-{\bf h}(G(\mathbf{z}^{l}))+\mathbf{J}^l \mathbf{z}^{l}) ,
\label{Gauss_Newton}
\end{align}
where $\mathbf{z}^{l}$ is the value of the latent space variable ${\bf z}$ at the $l$-th iteration and $\mathbf{J}^l$ is the Jacobian of the forward map (from the latent space to flow velocities) at the $l$th iteration (see below). The maximum of the total number of iterations (running time limit) as well as the minimum magnitude of the gradient (convergence of $\mathbf{z}$) can be considered as the stopping criteria during the Gauss-Newton iterations. 

The Jacobian $\mathbf{J}^l$ at the $l$th iteration is also given by 
\begin{equation}
    \mathbf{J}^l  = \left.\frac{\partial {\bf h}(G(\mathbf{z}))}{\partial \mathbf{z}}\right|_{\mathbf{z}= \mathbf{z}^l} = \left.\frac{\partial {\bf h}}{\partial \mathbf{s}}\right|_{\mathbf{s}=G(\mathbf{z}^l)}\left.\frac{\partial \mathbf{s}}{\partial \mathbf{z}}\right|_{\mathbf{z}= \mathbf{z}^l}  = \left.\left. \mathbf{J}_{\bf h} \right|_{\mathbf{s}=G(\mathbf{z}^l)} \mathbf{J}_G \right|_{\mathbf{z}= \mathbf{z}^l} .
\label{Jacob}
\end{equation}
Note that $\frac{\partial \mathbf{s}}{\partial \mathbf{z}}$ can be evaluated analytically using automatic differentiation (AD). This information can be provided at no additional computational cost in common libraries such as TensorFlow~\cite{van2017automatic} and PyTorch~\cite{paszke2017automatic}. Since the dimension of ${\bf z}$, $k \le 100$, is assumed to be significantly smaller than the dimension of ${\bf s}$, we can also use a finite difference formulation to calculate the Jacobian matrix as an alternative to AD: 
\begin{align}
\label{FD}
    \mathbf{J}^l \boldsymbol\Sigma (\mathbf{J}^l)^{\top} &= \mathbf{J}^l \boldsymbol\Sigma^{1/2} \left(\mathbf{J}^l \boldsymbol\Sigma^{1/2}\right)^{\top}\\
    \mathbf{J}^l \boldsymbol\Sigma^{1/2} &= \left[ \mathbf{J}^l \sigma_1 e_1, \mathbf{J}^l \sigma_2  e_2, \mathbf{J}^l \sigma_3 e_3, \ldots, \mathbf{J}^l \sigma_n e_n \right]\nonumber\\
    \mathbf{J}^l \sigma_i  e_i &\approx \sigma_i \frac{{\bf h}(G(\mathbf{z}+\delta_i)) - {\bf h}(G(\mathbf{z}))}{\delta_i} , \nonumber
\end{align}
where $e_i$ is a unit vector with $i$th element equal to 1 and $\delta_i$ is the finite difference discretization value. The number of forward model runs will be $k + 1$ at each iteration. Since the optimization problem in \Cref{z_map} becomes nonlinear because of the operator $G({\bf z})$, a line-search method has also been implemented. Also, for our numerical validation purposes, we assume throughout the remainder of the paper that $\boldsymbol\Sigma$ is the identity matrix.

Once the iterations are converged, for the uncertainty quantification (UQ) purposes, the posterior covariance of ${\bf z}$ can be calculated via 
\begin{align}
    \mathbf{Q}_\text{post} &= \left(\boldsymbol\Sigma^{-1} + \mathbf{J}^{\top}\mathbf{R}^{-1}\mathbf{J}\right)^{-1} \nonumber\\
    &= \boldsymbol\Sigma - \boldsymbol\Sigma \mathbf{J}^{\top} \left( \mathbf{J} \boldsymbol\Sigma {\bf J}^{\top} + \mathbf{R} \right) \mathbf{J} \boldsymbol\Sigma,
\label{Q_post}
\end{align}
where ${\bf J}$ is the Jacobian, calculated either through AD or \Cref{FD} at the converged ${\bf z}$.

\subsection{ROM}
\label{SVE_method}

In this section, we introduce the SVE architecture, the DNN-based ROM that we used in VEGAS. The purpose of this DNN is to find the map ${\bf s}= G({\bf z})$ from the low-dimensional latent space to the bathymetry and the map ${\bf h}(G({\bf z}))$ from the latent space to the flow velocities (see \Cref{z_map}). Due to the close connection between the SVEs and variational autoencoders (VAEs), we first provide a brief overview of the VAE structure and its purpose in deep learning problems. VAEs are a class of DNNs that are primarily used for nonlinear dimension reduction in unsupervised learning~\cite{Kramer, Hinton}. In VAE architectures, a high-dimensional input is fed as the input to the network; in the middle of the network (bottleneck), the dimension is reduced and two parallel layers provide the mean and variance of a multi-variate Gaussian distribution, from which the latent space variable is sampled; finally, the sampled latent space variable returns to the space of the output (which is the same as the input due to the unsupervised setup of VAE). This probabilistic structure imposes a strong regularization effect on the VAE architecture, which potentially leads to better generalization of the network~\cite{VAE_ref}. The network, in its unsupervised fashion, learns a map from a dataset to itself, the so-called auto-associative neural networks (NNs). However, the strength of the VAE is in its ability to simultaneously learn a map from the input to the bottleneck layer, referred to as the encoder, as well as a map from the bottleneck layer to the output, the decoder.

While VAE has been primarily introduced as a nonlinear dimension reduction technique in unsupervised learning (input and output being the same data), its input and output can be replaced with a set of labeled datasets (such as bathymetry and flow velocity), representing a supervised DNN architecture. Such architectures are very useful for problems in which the dimension of the input and output are very large, since in general, it is very difficult to build a DNN without reducing the dimension of the layers for such problems; otherwise, the number of trainable parameters becomes very large, which leads to a DNN that is computationally very expensive to train. Furthermore, such dense networks can lead to severe overfitting without a suitably large dataset.

\begin{figure}[htbp]
%\vspace{-1cm}
%\hspace{-1cm}
\centering
\begin{subfigure}{1.0\textwidth}
\centering
\includegraphics[width=1.0\linewidth]{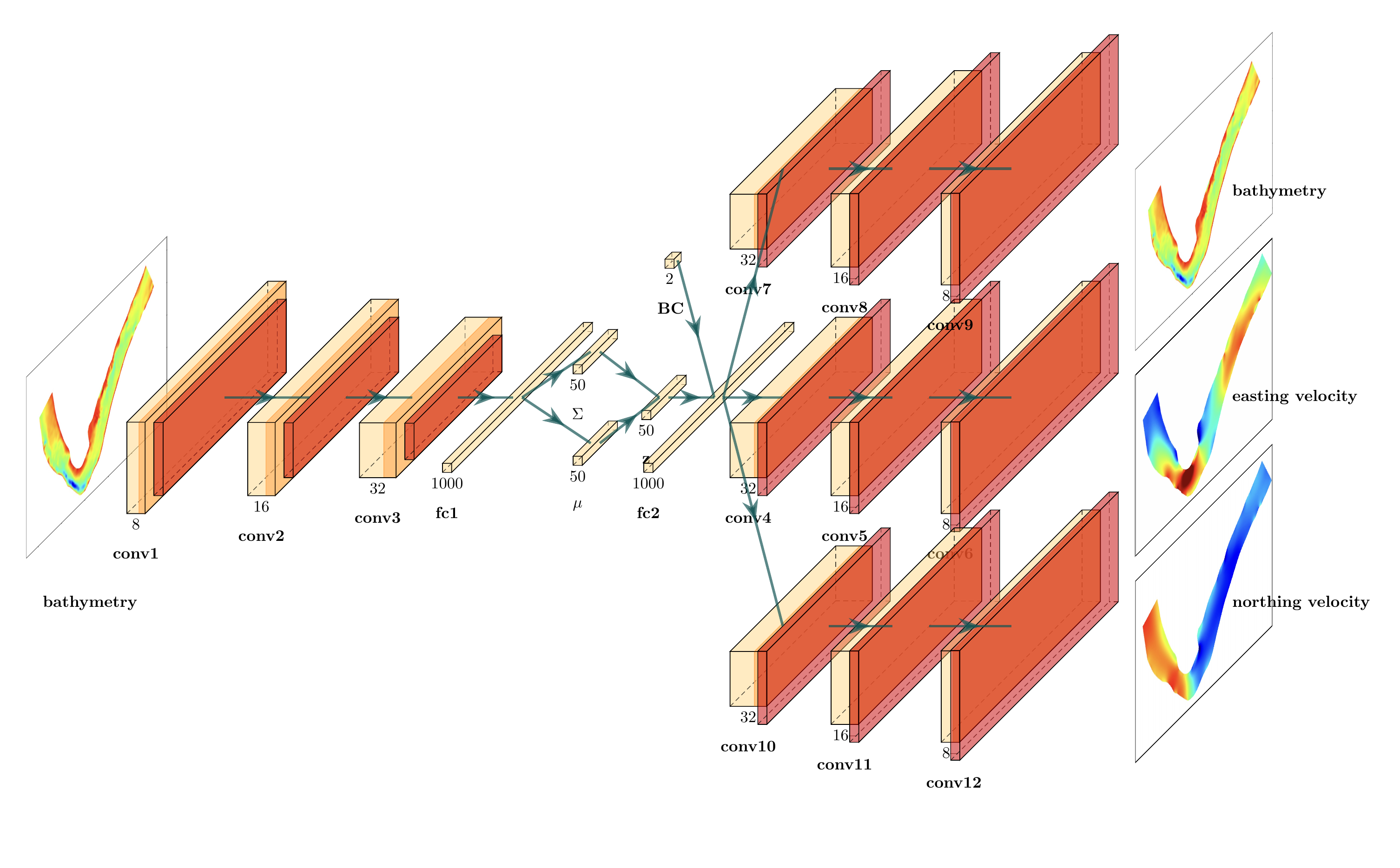}
\vspace{-1cm}
\caption{SVE}
\label{VAE_sketch}
\vspace{1cm}
\end{subfigure}
\begin{subfigure}{0.59\textwidth}
\centering
\includegraphics[width=1.05\linewidth]{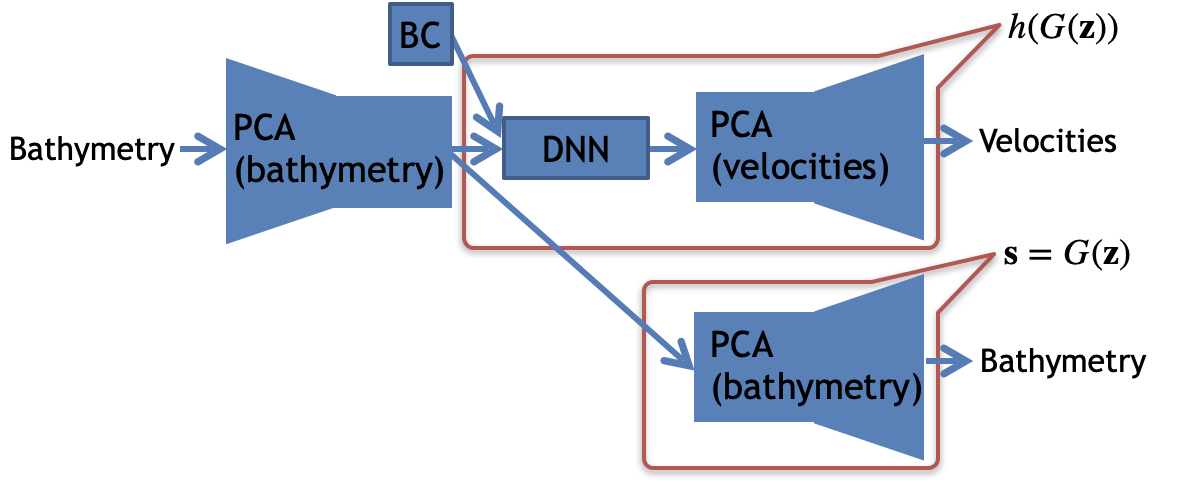}
\caption{PCA-based solver}
\label{PCA_sketch}
\end{subfigure}
\caption{(a) The schematic of the SVE architecture. In this approach, first, the dimension of the input (bathymetry) is reduced via an encoder, and then the low-dimensional data (the latent space) in the bottleneck layer is expanded to the output dimension. In SVE, the bottleneck layer generates a Gaussian distribution. The output of the decoder consists of both velocities and bathymetries. (b) The PCA-based solver replaces encoder/decoders with PCA, thus performing a linear dimension reduction/expansion.}
\label{VAE_PCA_sketch}
\end{figure}

In this work, we use convolutional layers in the SVE structure~\cite{Bishop} followed by a few fully connected (FC) layers that connect them to the bottleneck layer; this architecture is referred to as the convolutional SVE. Since in a convolutional network, filters are being applied to the whole input image (such as the 2D image of a riverbed profile), it reduces the size of the network significantly by taking advantage of the 2D nature of the input and homogeneity of the extracted features throughout the image. \Cref{VAE_sketch} shows the sketch of the SVE used in our work. The input in this figure is the bathymetry (the variable ${\bf s}$), while the output has three components, the flow velocity in two directions (function ${\bf h}(G({\bf z}))$) and the bathymetry (function ${\bf s}= G({\bf z})$). Note that the reason behind these outputs is that VEGAS requires the joint input/output distribution $\left( {\bf s}= {\bf h}(G({\bf z})), {\bf y}\right)$ (see \Cref{z_map}). More details on ${\bf h}(G({\bf z}))$ and $G({\bf z})$ and their relation to the inverse problem have been provided in Section \ref{bayes_meth}. ${\boldsymbol\mu}$ in this figure is the generated mean vector, $\boldsymbol\Sigma$ is the generated variance, and ${\cal N}(\boldsymbol\mu, \boldsymbol\Sigma)$ is the Gaussian distribution from which the latent variable ${\bf z}$ is generated (${\bf z}\sim {\cal N}(\boldsymbol\mu,\boldsymbol\Sigma)$). We have also used a linear dimension reduction-based forward solver for comparison purposes in our work, in which case the dimension reduction and expansion are performed linearly via a PCA instead of nonlinearly using encoder and decoders; this is shown in \Cref{PCA_sketch}.

\section{Study site and data}
\label{data_prep}

In this section, we discuss the process used to obtain the training data for our DNN (the architectures in \Cref{VAE_PCA_sketch}). Here, we are assuming that a sparse flow velocity measurement of the river of interest is available, and thus, we generate a training dataset that is relevant to this observation. Note that the VEGAS does not require the existence of such measurements at the time of the training process (offline stage). However, the purpose of this section is to propose an efficient way of generating a training dataset that takes such information (if available) into account, in order to generate a dataset that is relevant to our historical observation.

\begin{figure}[htbp]
%\vspace{-1cm}
%\hspace{-1.5cm}
\centering
\begin{subfigure}{.5\textwidth}
\centering
\includegraphics[width=1.0\linewidth]{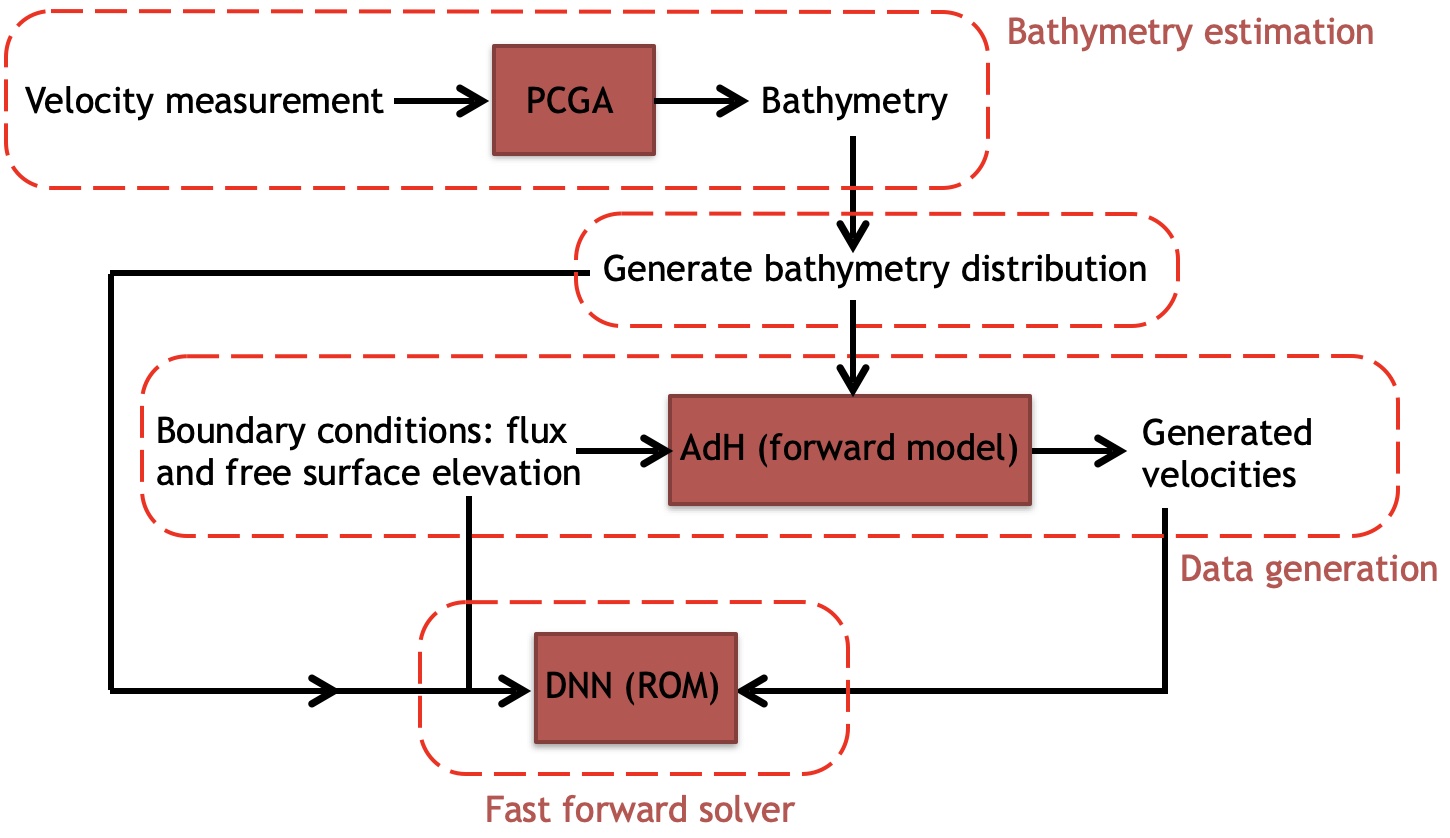}
\caption{}
\label{sketch}
\end{subfigure}
\begin{subfigure}{.4\textwidth}
\centering
\includegraphics[width=1.1\linewidth]{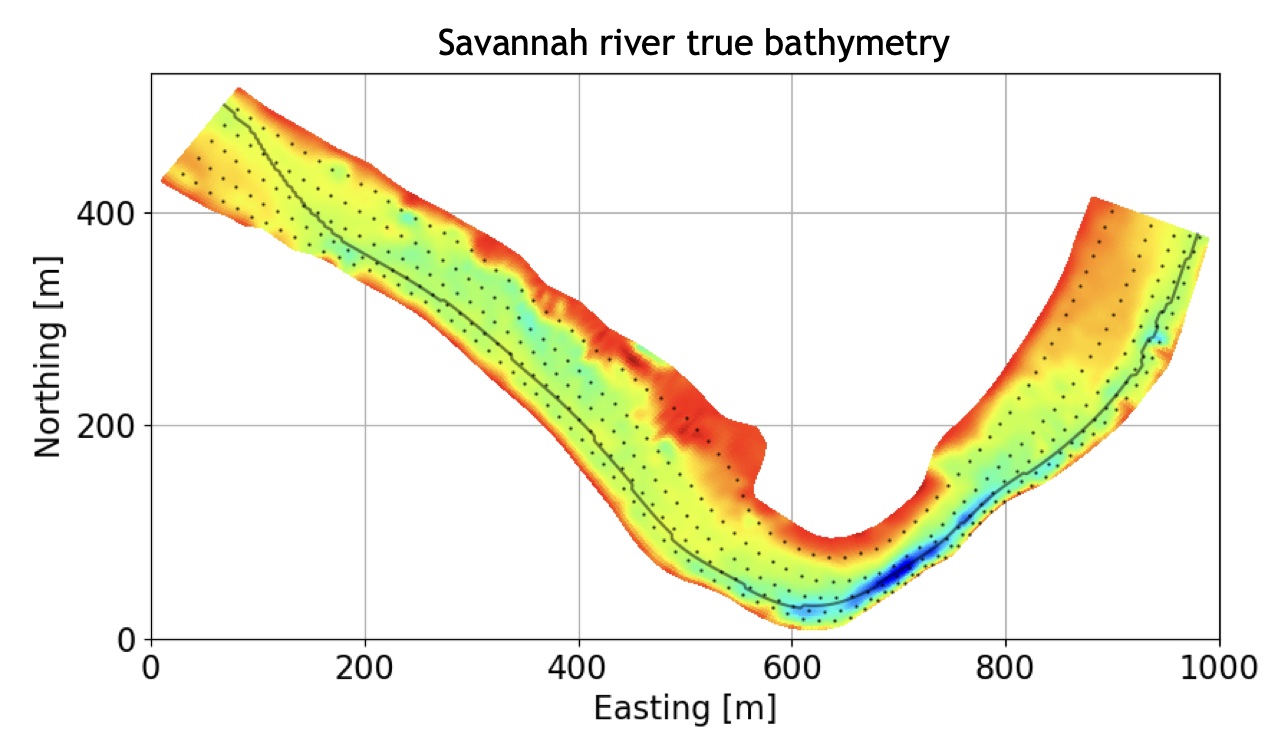}
\vspace{0.0cm}
\caption{}
\label{vel_loc}
\end{subfigure}
%\vspace{-2cm}
\caption{(a) Schematic of the development of the DNN-based ROM. First, we estimate the posterior distribution of the bathymetry via the PCGA, then augment this distribution to a more general distribution in order to allow the solver to include a larger class of bathymetries into its prediction capability, and then use AdH to generate velocities. Finally, the bathymetries, BCs, and velocities are fed to the SVE which will be used as the ROM. (b) Reference (true) bathymetry of the Savannah river. The black line shows location of the thalweg, the deepest point along the river, at a given cross section. The dots are the 408 measurement locations used as inputs to the PCGA.}
\label{data_steps}
\end{figure}

\Cref{sketch} shows the steps being used for generation of the dataset (training/validation/test) necessary for learning the ROM (the SVE network). At the first stage, an estimation of the bathymetry of the river is obtained from the sparse flow velocity measurement via PCGA~\cite{PCGA_Kitanidis, PCGA_Lee}. In this case, the river flow velocity measurement is performed during a short-term data collection campaign with a fixed bathymetry assumption. In order to allow our DNN to include a larger class of bathymetries into its prediction capability, we generate a bathymetry distribution from the estimated bathymetry of the PCGA (the ``generate bathymetry distribution'' step in \Cref{sketch}) via an additional sampling process based on typical river topography, before feeding them as inputs to the DNN (the SVE architecture in \Cref{VAE_sketch}). This allows the solver to be accurate for flow velocity prediction when future bathymetries, different from the posterior estimate, are provided as new inputs. The generated distribution along with the BCs are fed to the high-fidelity solver (the ``AdH'' block), which generates the data necessary for training the SVE (the ``DNN'' block). More detail of this process can be found in \cite{SWE_forward}.

In this work, the VEGAS has been validated on a roughly one mile reach of the Savannah river near Augusta, GA (see \Cref{vel_loc}). First, we use AdH to generate (sparse) noisy flow velocity measurements synthetically, and then apply PCGA to these measurements. The PCGA-estimated bathymetries are then augmented via a Gaussian kernel before providing them as inputs to the AdH. The augmentation ensures that a larger class of bathymetries (larger standard deviation than the posterior estimate of the PCGA) are included in the training set. These bathymetries consist of $41\times 501= 20,541$ mesh nodes (501 nodes in the along-channel and 41 nodes in across-channel direction). Finally, these bathymetries along with the generated BCs---here assumed to be discharge and free-surface elevation (extracted from the United States Geological Survey (USGS) website~\cite{USGS_web}) are provided as inputs to AdH in order to obtain the flow velocities. These bathymetry/BC/flow velocity datasets are the training dataset that will be fed to the DNN in order to obtain the ROM (the SVE of \Cref{VAE_sketch}).

\section{Results and discussion}
\label{result_global}

In this section, we provide the result of applying our inversion approach to the dataset generated in Section \ref{data_prep}. In Section \ref{global_full}, we provide the result of inversion (bathymetry reconstruction) on the dataset of Section \ref{data_prep} when dense flow velocity measurements are available. We also compare the performance of VEGAS with a similar inversion approach whose DNN-based ROM uses linear dimension reduction technique~\cite{SWE_forward} (see \Cref{PCA_sketch}). In Section \ref{global_partial}, we provide similar results in the presence of sparse flow velocity measurement. And finally in Section \ref{unseen_distb}, we present the inversion result on a test dataset that is sampled from a distribution different from the one that the SVE is trained on.

\subsection{Performance of VEGAS in the presence of full measurement}
\label{global_full}

In this section, we show the result of applying the inversion algorithm to the Savannah river domain. Here, we are assuming that the dense flow velocity measurement is available to the solver, therefore, dim(${\bf y}$)$=20,541$ ($=41\times501$) for flow velocity in each direction. We will discuss the results in the presence of the sparse flow velocity measurement in Section \ref{global_partial}. In Section \ref{latent}, we study the influence of the dimension of the latent space on the inversion process and its ability to find a low-dimensional representation of the SWE dynamics.

\begin{figure}[htbp]
\centering
\begin{subfigure}{1.0\textwidth}
\centering
\includegraphics[width=1.0\linewidth]{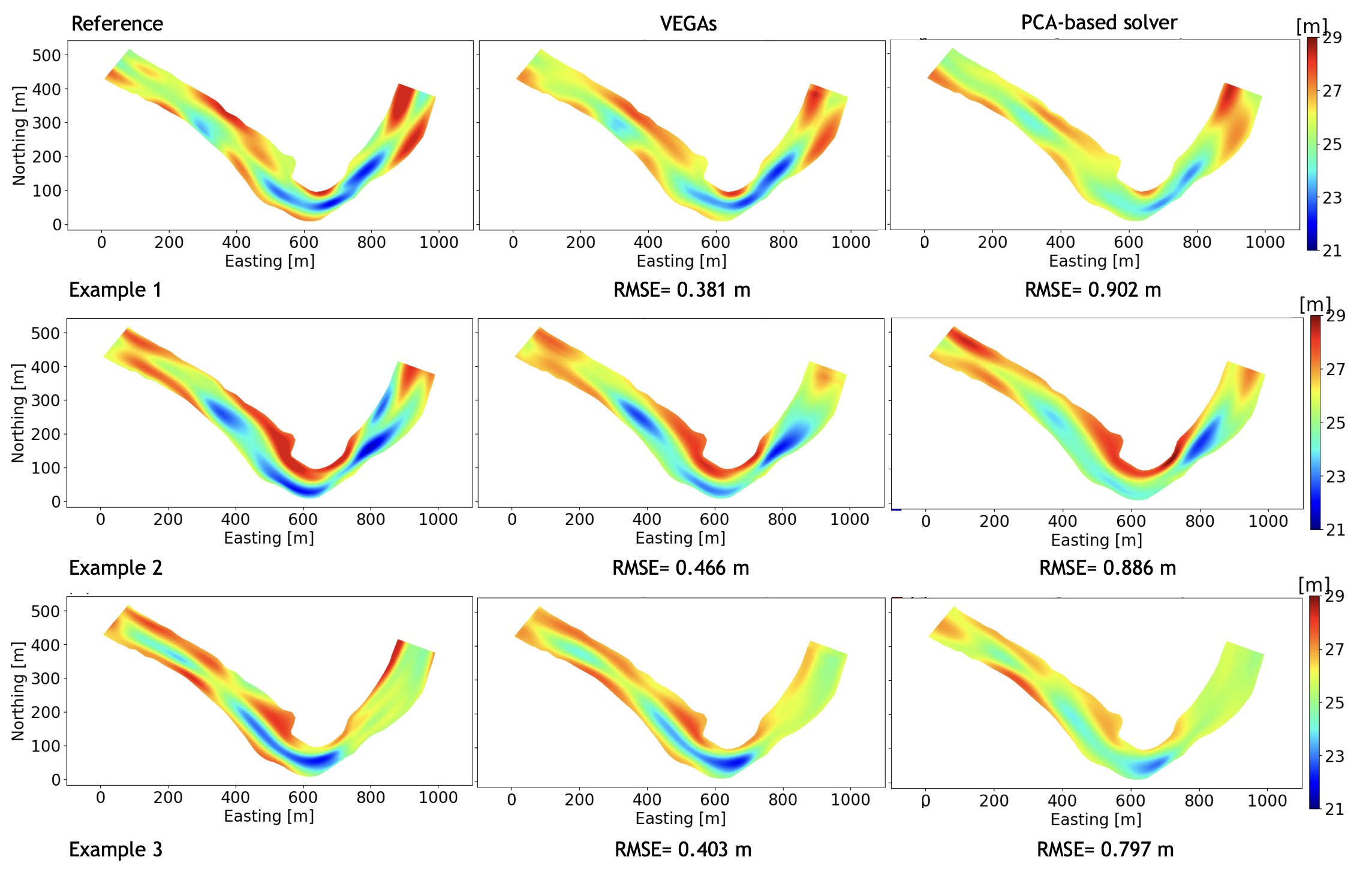}
%\caption{}
\label{ab_initiomodel}
\end{subfigure}
\caption{Comparison between the performance of VEGAS and the PCA-based solver for the bathymetry for three different test cases. Superior performance of VEGAS is noticeable in all cases. In both approaches, we chose a latent space dimension of dim$({\bf z})= 50$.}
\label{plots_inv_ex}
\end{figure}

\begin{figure}[htbp]
\centering
\begin{subfigure}{1.0\textwidth}
\centering
\includegraphics[width=0.70\linewidth]{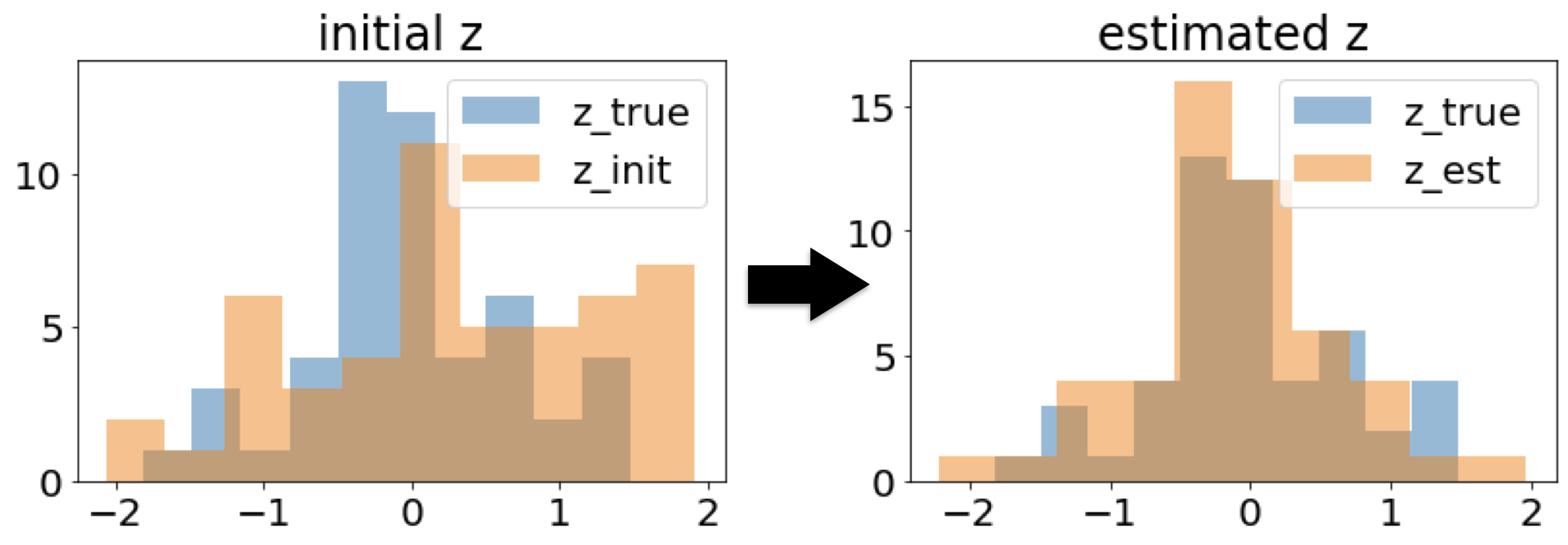}
%\caption{}
\label{ab_initiomodel}
\end{subfigure}
\caption{Initial versus final distribution of the latent space variable for a test set data point during the inversion process of VEGAS. The final estimated latent space variable matches the true distribution.}
\label{plots_z_est}
\end{figure}

Table \ref{error_inv} summarizes the root mean square errors (RMSEs) when estimating the bathymetries from the flow velocity measurements (inverse problem) using different methods (SVE and the PCA-based ROMs) in the presence of dense measurements. Here, we used 10$\%$ of our data for validation and a different set of unused 450 input-output pairs for testing. In order to have a fair comparison between different methods, we used the same latent space dimension in both methods, equal to 50 (see Section \ref{latent} for further detail regarding this choice). The errors for VEGAS are significantly lower than the PCA-based solver, due to the linear dimension reduction technique being used in the PCA-based approach. The inversion iterations with a preset maximum number of iterations of 10 takes about 1--3 min on an Intel(R) Xeon(R) CPU E5-2699 v3 @ 2.30 GHz machine with 8 cores. \Cref{plots_inv_ex} compares performance of VEGAS with the PCA-based solver for three data points from the test set. We observe that in all cases VEGAS performs significantly better than the PCA-based approach, consistent with the result of table \ref{error_inv}. \Cref{plots_z_est} shows an example of the reference (true) and reconstructed latent space distribution for a test data point. The figure compares the initial and final distribution of ${\bf z}$ with its true distribution. We observe that the algorithm has been successful in capturing the true distribution, starting from an initial random distribution.

\begin{table}[htbp]
    \centering
    \begin{tabular}{lllll}
        \toprule
        \multirow{2}{*}{Inversion method} & \multicolumn{3}{c}{Error (RMSE)}\\
        \cmidrule{2-4} \cmidrule{5-5}
        {} & training set [m] & validation set [m] & test set [m] \\
        \midrule
        VEGAS   & {\bf 0.397} & {\bf 0.413} & {\bf 0.461}\\
        PCA-based method & 0.818 & 0.794 & 0.834\\
        \bottomrule
    \end{tabular}
    \caption{Comparison between the reconstruction error of different inversion methods.}
\label{error_inv}
\end{table}

\begin{figure}[htbp]
\centering
\hspace{-0.45cm}
\begin{subfigure}{1.0\textwidth}
\centering
\includegraphics[width=0.85\linewidth]{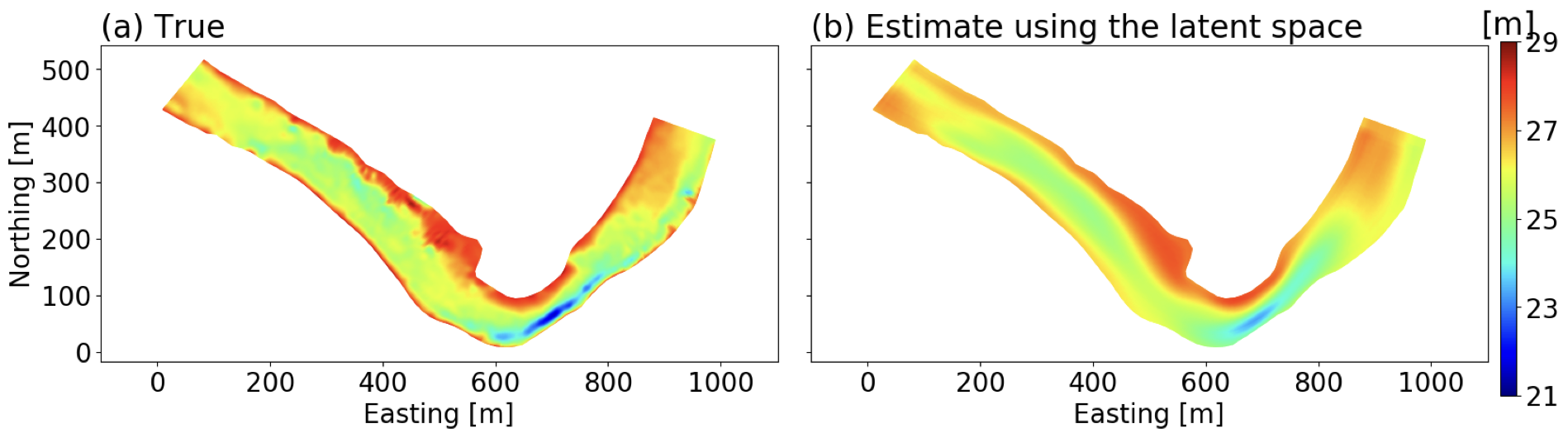}
%\caption{}
\label{error_dist}
\end{subfigure}
\hspace{0.2cm}
\caption{Reconstruction of the Savannah river profile.}
\label{sav_res}
\end{figure}

We have also evaluated the performance of the inversion algorithm on the original reference profile for the Savannah river reach obtained from the U.S. Army Corps of Engineers survey. The result is shown in \Cref{sav_res}. We observe that the algorithm has been able to reconstruct the true profile with good accuracy. Here, we have assumed that the full flow velocity measurements are available to the inversion algorithm.

\subsubsection{Latent space dimension}
\label{latent}
\begin{figure}[htbp]
\centering
\begin{subfigure}{1.0\textwidth}
\centering
\includegraphics[width=1.0\linewidth]{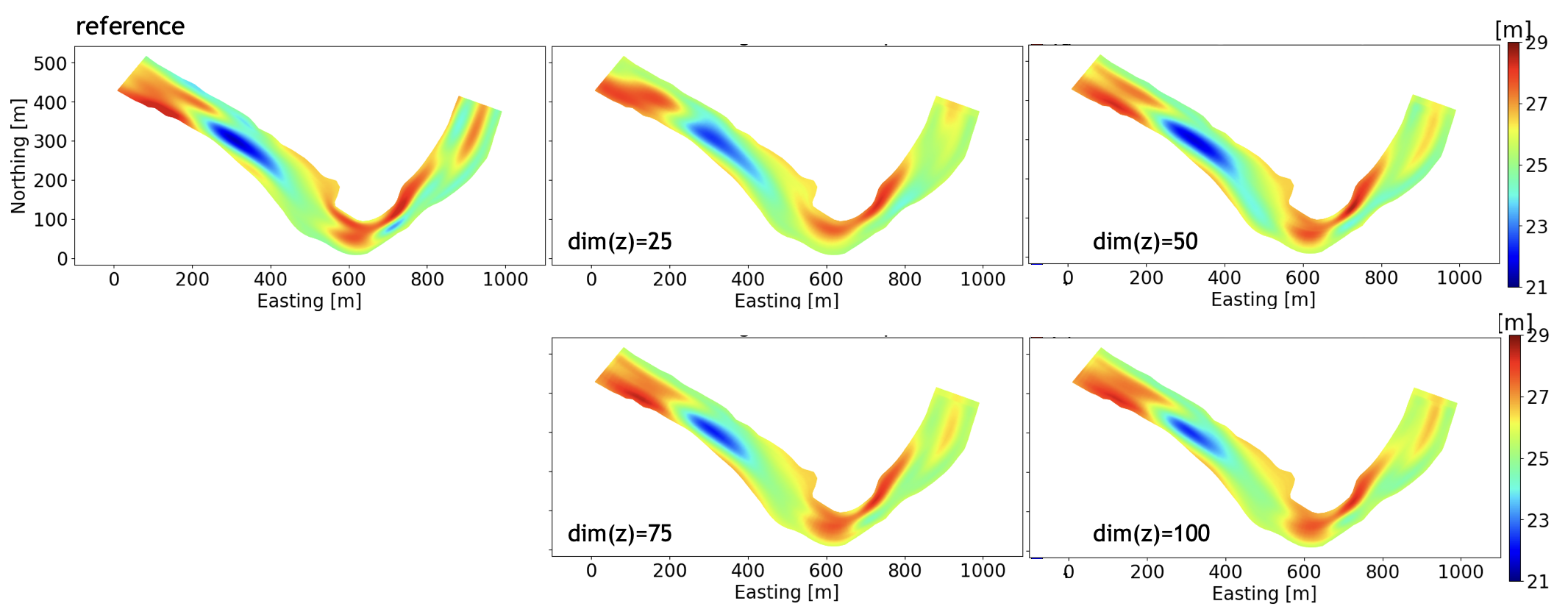}
\caption{}
\label{ab_initiomodel}
\end{subfigure}
\begin{subfigure}{1.0\textwidth}
\centering
\includegraphics[width=1.0\linewidth]{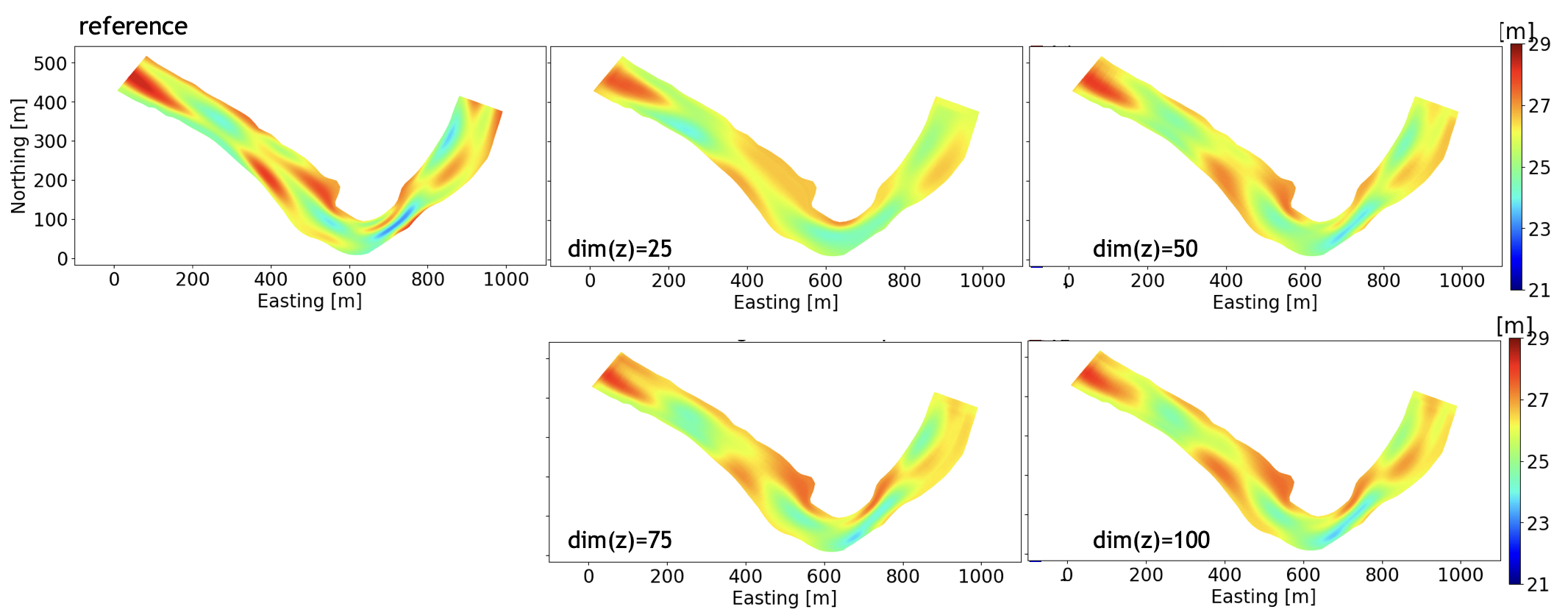}
\caption{}
\label{Hollandmodel}
\end{subfigure}
\caption{Examples of the bathymetry reconstruction on two test set data points for different latent space dimensions.}
\label{plots_dim}
\end{figure}

In this section, we study the effect of dimension of the latent space on the quality of the reconstruction and whether the dimension used in previous sections (dim$({\bf z})=50$) is sufficiently large to capture the dynamics of the forward/inversion problem. \Cref{plots_dim} shows examples of the bathymetry reconstruction for two different test set data points for different latent space dimensions (dim$({\bf z})=25, 50, 75, 100$). We observe that in general the quality of the reconstruction improves as we increase the latent space dimension. However, the improvement observed in the case of dim(${\bf z}$)$=50$ compared to dim(${\bf z}$)$=25$ is more significant. This trend is noticeable in the other test data points as well.

\begin{figure}[htbp]
\centering
%\hspace{-1.2cm}
\begin{subfigure}{1.0\textwidth}
\centering
\includegraphics[width=0.80\linewidth]{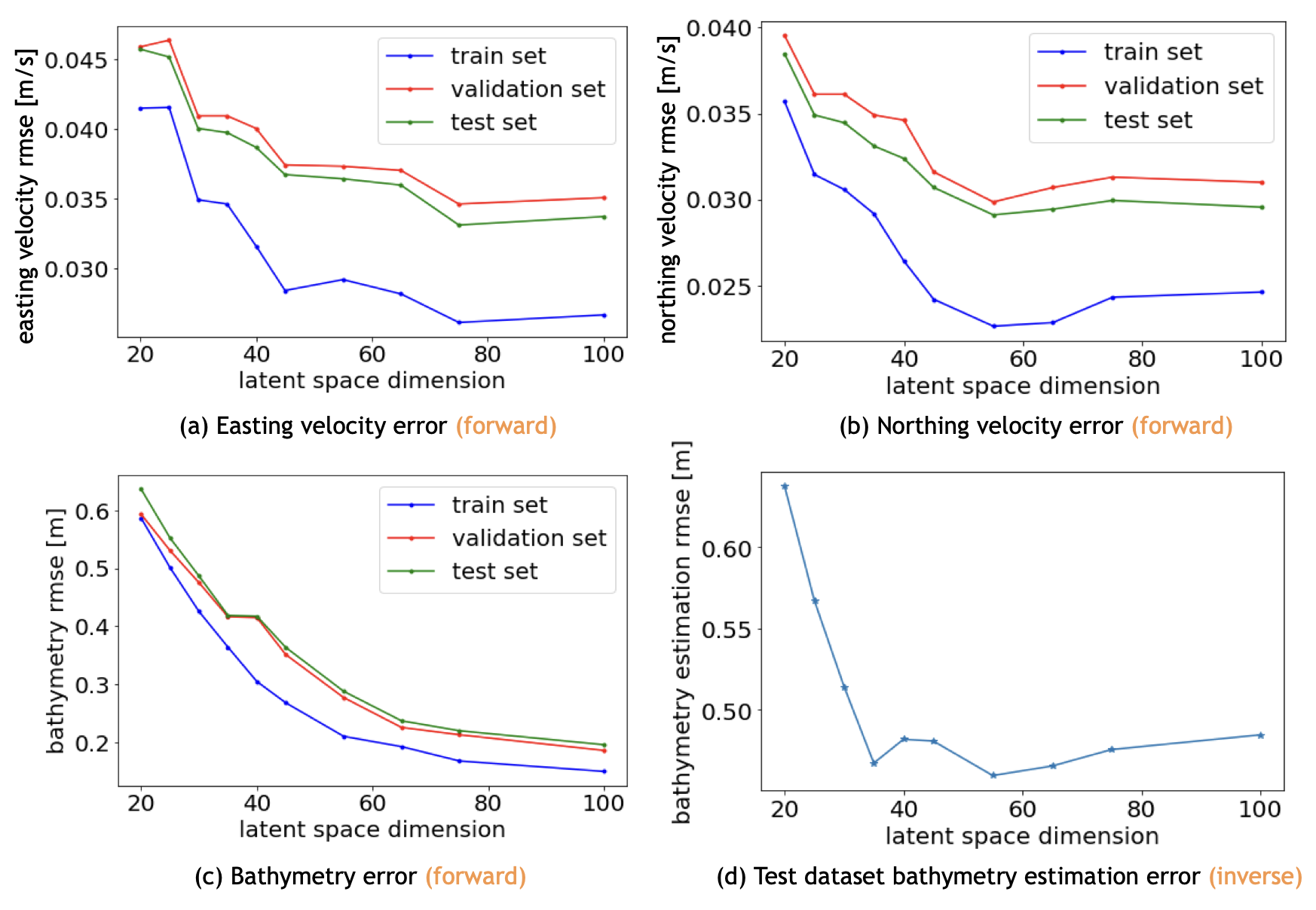}
%\caption{}
\label{ab_initiomodel}
\end{subfigure}
\caption{RMSE of the SVE prediction for different latent space dimensions for the flow velocity in the easting direction (a), northing direction (b), and the bathymetry (c). And RMSE of the inversion (VEGAS) on the test dataset (d). Although the errors decrease past dim(${\bf z}$)$=50$ in the case of bathymetry (subfigure (c)), no significant improvement in the VEGAS performance is observed after this point.}
\label{errs_dim}
\end{figure}

In order to better quantify the effect of latent space dimension on the reconstruction quality and, more importantly, determine the correct number of latent space dimensions to capture the dynamics of inverse and forward solvers, we also plot the RMSE of the training, validation, and test datasets for the ROM (the SVE), as well as the RMSE of the test dataset for the inversion algorithm (the VEGAS), as a function of the latent space dimension. \Cref{errs_dim} shows these results. We can observe that for the SVE, the flow velocities RMSE has stopped improving at dimensions in the range 50--70 (subfigures (a) and (b)). This dimension is found to be higher for the bathymetry (subfigure (c)), implying that a larger latent space dimension is required to represent the bathymetry accurately. However, almost no improvement is observed in the quality of the inversion problem (bathymetry reconstruction) after dimensions in the range 50--60 (subfigure (d)). This could likely be due to the fact that the VEGAS is based on, first, finding the low-dimensional representation of the velocity measurements (related to subfigures (a) and (b)), and then, expanding this representation to the bathymetry space (subfigure (c)). Therefore, although the bathymetry ROM improves past dim$({\bf z})=50$, its influence on the reconstruction quality is not very significant, since the accuracy of the estimated ${\bf z}$ does not improve at larger latent space dimensions. These observations imply that the chosen latent space dimension (dim$({\bf z})=50$) is sufficiently large to capture the dynamics of the forward/inverse problem for the training/validation/test sets that we have considered in this work. This is also consistent with the result of \Cref{plots_dim} (improvement after dim$({\bf z})=50$ is negligible).

\begin{figure}[htbp]
\centering
%\hspace{-1.2cm}
\begin{subfigure}{1.0\textwidth}
\centering
\includegraphics[width=0.95\linewidth]{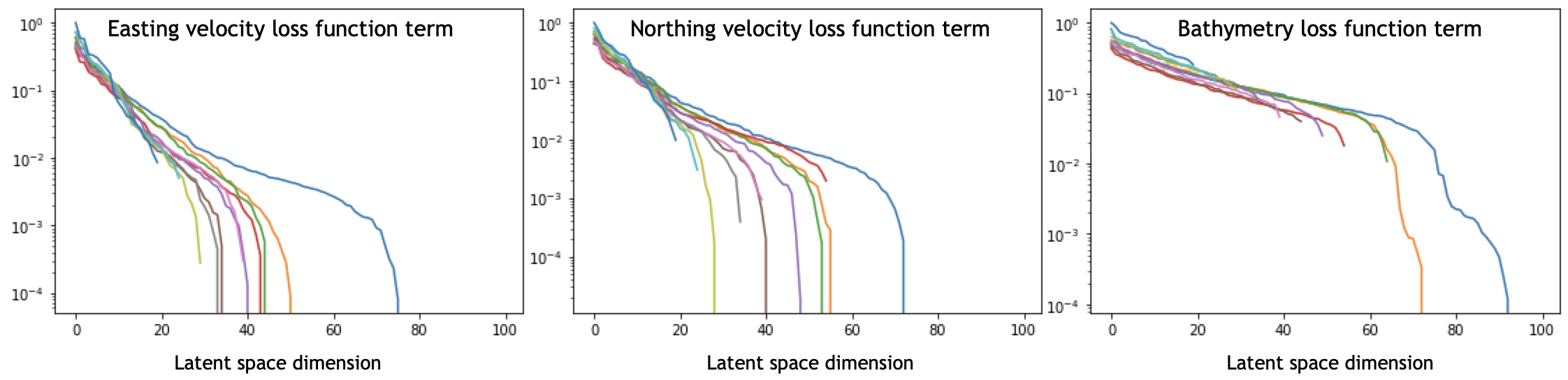}
%\caption{}
\label{ab_initiomodel}
\end{subfigure}
\caption{Eigendecomposition of the Hessian matrix for different terms of the loss function of the SVE with respect to the latent space variable. Each (colored) curve represents a SVE with a particular latent space dimension in the range dim(${\bf z}$)= 20--100. Decomposition of the easting velocity loss term (a), northing velocity loss term (b), and bathymetry loss term (c) for different latent space dimensions. Reaching zero values at smaller latent space dimensions for the flow velocities implies that they require fewer number of latent space dimension compared to the bathymetry.}
\label{hess_plot}
\end{figure}

We have also plotted the eigendecomposition of the Hessian of the loss function with respect to the latent space components to better understand the effect of the latent space dimension. When the Hessian matrix eigenvalues reach small values, it implies the existence of directions that are not influential in the dynamics. The results are shown in \Cref{hess_plot} for the ROM (SVE) for different latent space dimensions. The plots show the eigendecompositions of the Hessian, obtained via applying singular value decomposition (SVD) to the Hessian of the different terms of the loss function, that is, velocity in easting direction, velocity in northing direction, and bathymetry, calculated with respect to the components of the latent space variable ${\bf z}$. Note that the loss function of the DNN (the SVE) has three terms (\Cref{VAE_sketch}), corresponding to its three outputs (the two flow velocities and the bathymetry). The Hessian matrices are calculated numerically using a finite difference method. We observe that for the flow velocities, the Hessian reaches zero at smaller latent space dimensions, in particular, 35 for the easting velocity and 30 for northing velocity, while for the bathymetry it happens at larger latent space dimensions, 75. This is also consistent with our observation in \Cref{plots_dim} and \Cref{errs_dim}. Reaching zero values at smaller latent space dimensions for the flow velocities implies that they require smaller number of latent space dimension compared to the bathymetry.

\subsection{Performance in the presence of sparse measurements}
\label{global_partial}

\begin{figure}[htbp]
\centering
\begin{subfigure}{0.9\textwidth}
\centering
\includegraphics[width=1.0\linewidth]{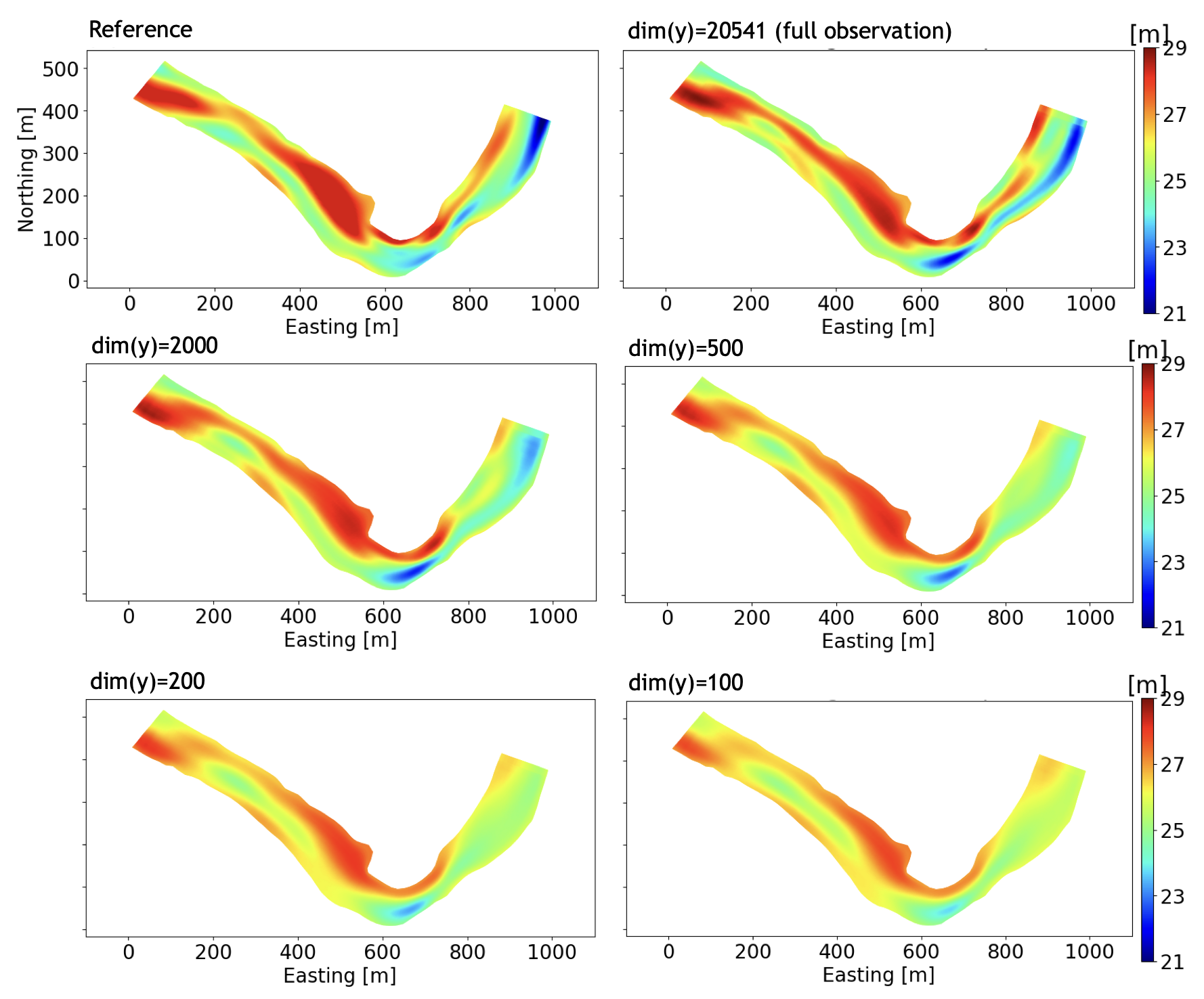}
%\caption{}
\label{ab_initiomodel}
\end{subfigure}
\caption{The effect of sparsity in flow velocity measurement on the reconstruction quality for a data point in the test dataset. The error increases as we use fewer velocity measurements $\bf y$ (see Section \ref{bayes_meth} for details). However, in all cases, the algorithm was mostly successful in capturing important features of the bathymetry, such as the locations with the largest and smallest depth.}
\label{sparse_dim}
\end{figure}

\begin{figure}[htbp]
\centering
\begin{subfigure}{0.45\textwidth}
\centering
\includegraphics[width=1.0\linewidth]{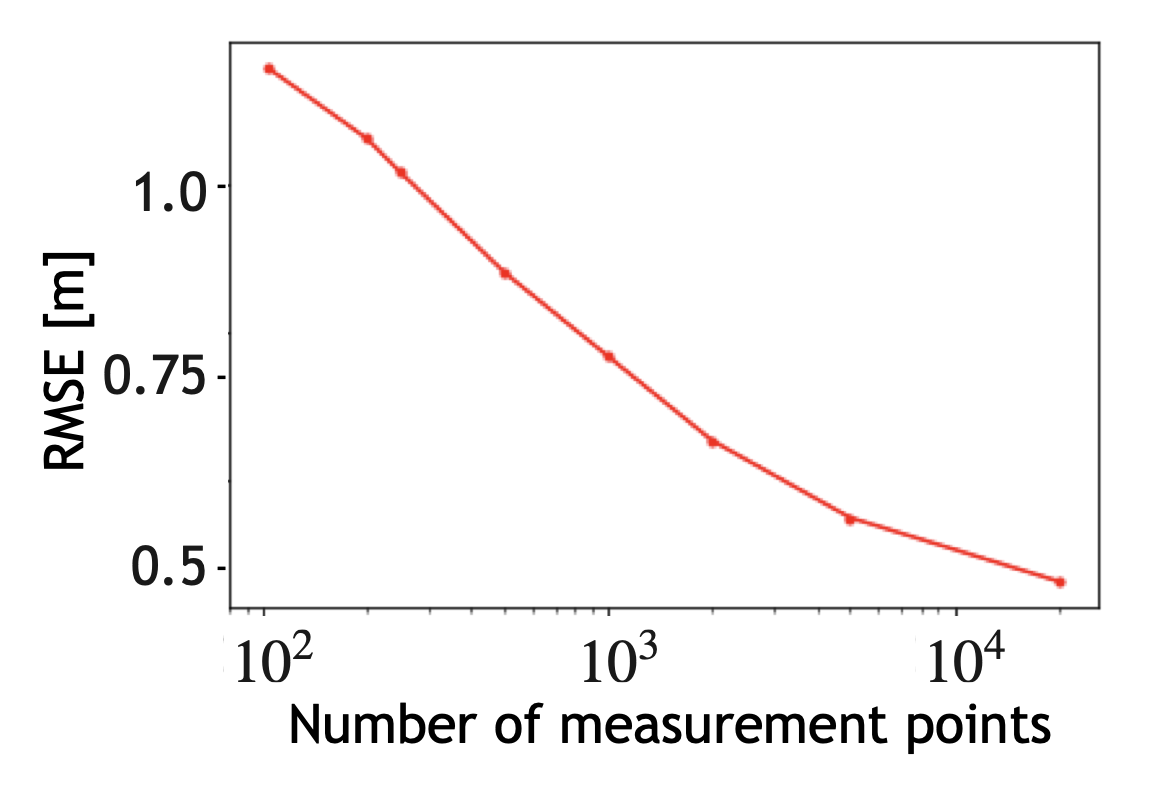}
%\caption{}
\label{ab_initiomodel}
\end{subfigure}
\caption{The effect of sparsity in the flow velocity measurements on the reconstruction quality for the test dataset. % The error increases as we use fewer measurements. However, the error remains reasonable even with significantly sparse measurements.
}
\label{sparse_err}
\end{figure}

The reconstruction results that we have shown so far assumed that full flow velocity measurements are available to the inversion algorithm, that is, the assumption is that $41\times501\times2$ measurements are provided to the VEGAS, where 41 and 501 are the number of grid points in the northing and easting directions and 2 accounts for flow velocities in the easting and northing directions. \Cref{sparse_dim} shows an example of a data point from the test dataset with its VEGAS prediction when full bathymetry measurement (20,541 for flow velocity in each direction), 2,000 measurement points, 500 measurement points, 200 measurement points, and 100 measurement points are available to the algorithm. The measurement points are equally distanced along and across the river (see \Cref{vel_loc} for an example of data points location). We observe that, as expected, the error increases as we use a smaller number of measurements. However, the error even with significantly sparse measurement is reasonable. \Cref{sparse_err} shows a similar comparison for the RMSE of the test dataset. Although the error increases as we use less number of measurements, we observe that even using very sparse measurements, such as the leftmost point which uses only 0.5\% of the measurements (100 measurement points versus 20,541 measurement points), the reconstruction error is reasonable. This is an important property of VEGAS, since in many practical applications, flow velocity measurements only at a limited number of locations are available and therefore, accurate reconstruction of the bathymetry using such sparse measurements is an important feature of any inversion algorithm.

\subsection{Performance on unseen test set distribution}
\label{unseen_distb}

The test and training datasets that we have used so far were sampled from the same distribution, the distribution explained in Section \ref{data_prep}. However, as it was shown in \Cref{sketch}, our distribution was obtained via augmenting the PCGA estimated bathymetry of the Savannah river. In this section, we are interested in evaluating the performance of VEGAS on datasets that are coming from a distribution that is different from the distribution that the DNN was trained on. 

In this section, we generate our training dataset by adding \textbf{a Gaussian kernel to a parabolic mean bathymetry} and test the performance of VEGAS trained with this distribution, on the Savannah river as well as our previously used dataset (PCGA posterior with added Gaussian kernel). Note that the advantage of this approach is that the training set does not require any initial flow velocity measurement. 

\begin{figure}[htbp]
%\vspace{-1cm}
\centering
%\hspace{-1.2cm}
\begin{subfigure}{1.0\textwidth}
%\vspace{-4cm}
\centering
\includegraphics[width=0.90\linewidth]{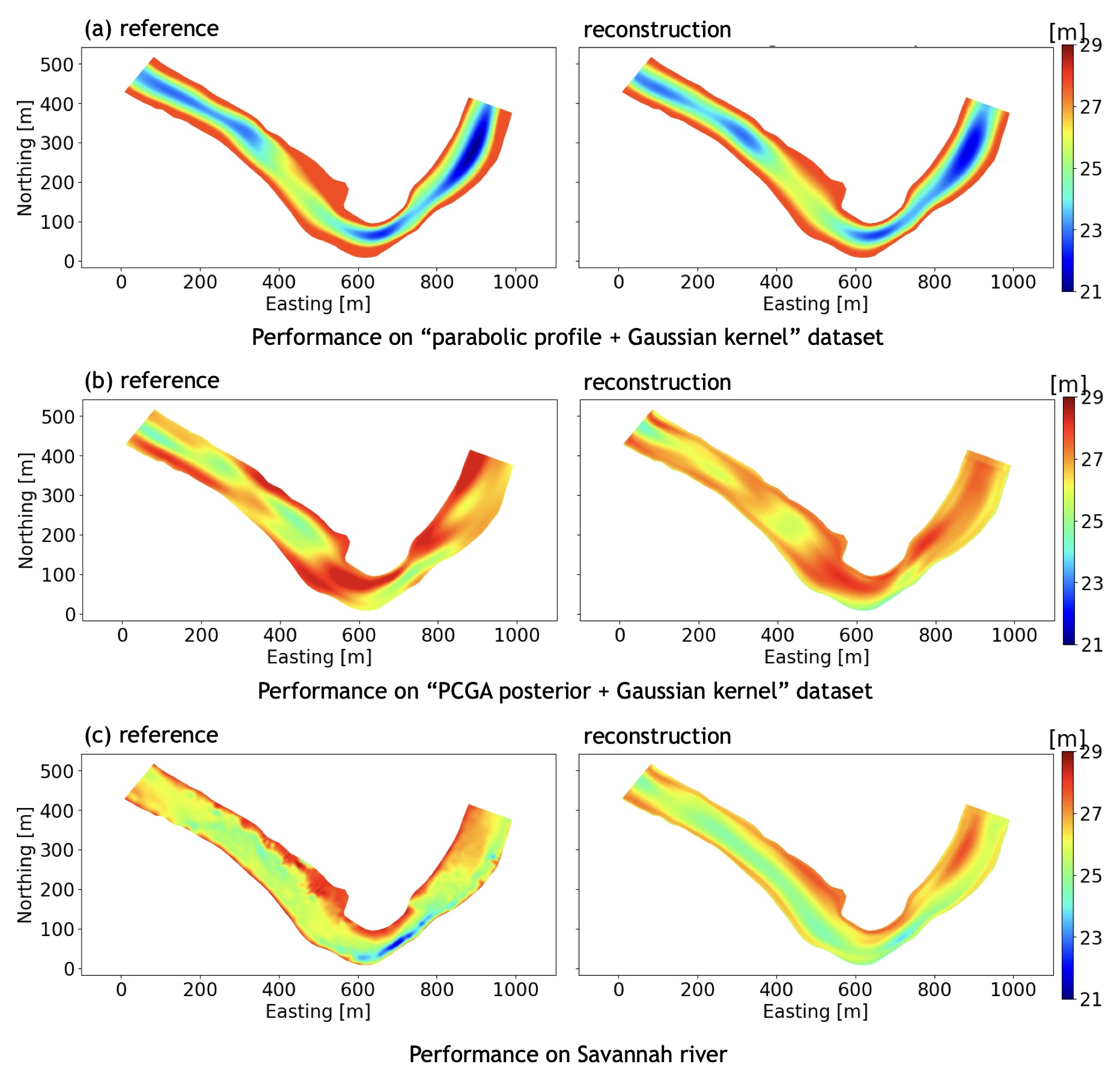}
%\caption{}
\label{d_local}
\end{subfigure}
\caption{Performance of VEGAS trained on ``parabolic +  Gaussian kernel'' and tested on (a) a data point from the same distribution, (b) a data point from ``PCGA estimated posterior mean +  Gaussian kernel'' distribution, (c) Savannah river.}
\label{unseen_dist}
\end{figure}

\begin{figure}[htbp]
%\vspace{-1cm}
\centering
%\hspace{-1.2cm}
\begin{subfigure}{1.0\textwidth}
%\vspace{-4cm}
\centering
\includegraphics[width=0.90\linewidth]{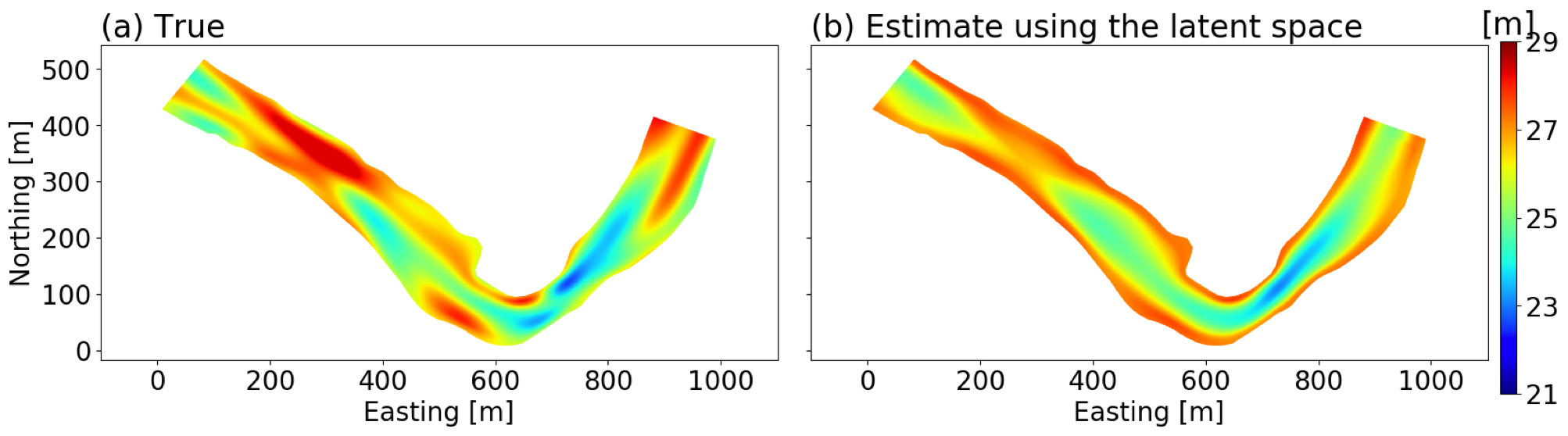}
%\caption{}
%\label{d_local}
\end{subfigure}
\caption{Performance of VEGAS trained on ``parabolic +  Gaussian kernel'' and tested on a data point from ``PCGA estimated posterior mean +  Gaussian kernel'' distribution with \textit{sparse flow velocity measurements} (10\% of measurements).}
\label{unseen_sparse}
\end{figure}

The results for different test set data points are shown in \Cref{unseen_dist}. The top figure shows the reconstruction on a test data point that was sampled from the same distribution that the training set was based on, parabolic with added Gaussian kernel. As expected, the error in this case is very low, since the test data point is sampled from the same distribution that the training set was based on. In the middle figure (\Cref{unseen_dist}b), the error is higher, since this data point is sampled from the previously mentioned dataset (PCGA posterior with added Gaussian kernel) which is different from the training set distribution. Finally, we show the reconstruction performance on the Savannah river. We observe that even in cases where the test data points are sampled from a distribution different from the training set distribution, the algorithm is capable of providing reasonably accurate reconstructions. The RMSE of the test set for the ``parabolic +  Gaussian kernel'' distribution is 0.48 m and for the ``PCGA posterior +  Gaussian kernel'' is 1.2 m. The larger RMSE of the latter is due to the fact that this test set is sampled from a distribution that is different from the training set distribution. \Cref{unseen_sparse} also shows an example of the reconstruction on a test data point from ``PCGA posterior +  Gaussian kernel'' distribution (similar to \Cref{unseen_dist}b) with sparse flow velocity measurements (2,000 measurement points). The quality has deteriorated, as expected. However, the reconstruction was able to extract the most important features of the profile. 

\begin{figure}[htbp]
\centering
\begin{subfigure}{1.0\textwidth}
\centering
\includegraphics[width=1.0\linewidth]{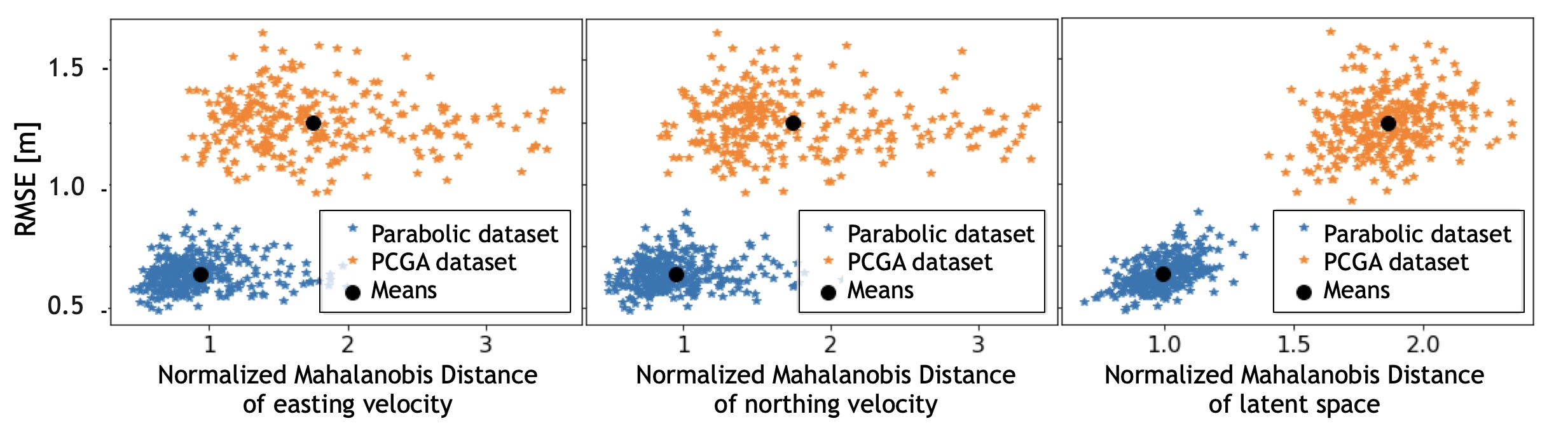}
%\caption{}
\label{ab_initiomodel}
\end{subfigure}
\caption{Comparison between the RMSE of the reconstructed bathymetries using VEGAS with the test set ``parabolic +  Gaussian kernel'' (``Parabolic dataset'' label in the figure) and ``PCGA posterior +  Gaussian kernel'' distribution (``PCGA dataset'' label in the figure) as a function of the Mahalanobis distances for easting velocity (left), northing velocity (middle), and latent space (c). ``Parabolic +  Gaussian kernel'' is the training set for this figure.% 
%The ratio between the means of the RMSEs and distances are approximately preserved for the two clusters.
}
\label{mahal}
\end{figure}

Although the errors in the cases that the test set distribution is different from the training set is higher than the cases that these two distributions are the same, as seen in \Cref{unseen_dist}, this larger error is expected as the two distributions become farther from each other. Here, we are quantifying the effect of the distance between the training and test set distributions on the test set error. We are using the normalized Mahalanobis distance for this purpose, defined as
\begin{equation}
    d=\sqrt{\frac{({\bf x} - \boldsymbol\mu_\text{train})^\top\boldsymbol\Sigma_\text{train}^{-1}({\bf x}- \boldsymbol\mu_\text{train})}{m}} ,
\label{maha_def}
\end{equation}
where ${\bf x}$ is the sample test data point, $\boldsymbol\mu_\text{train}$ is the mean of the training set distribution, $\boldsymbol\Sigma_\text{train}$ is its covariance matrix, and $m$ is the dimension of the dataset. For the analyses considered here, $\boldsymbol\mu_\text{train}$ and $\boldsymbol\Sigma_\text{train}$ are the distribution properties of the ``parabolic + Gaussian kernel'' training set, while ${\bf x}$ can belong to either ``parabolic + Gaussian kernel'' (\textit{e.g.}, \Cref{unseen_dist}a) or ``PCGA posterior + Gaussian kernel'' (\textit{e.g.}, \Cref{unseen_dist}b) distribution. We have plotted the RMSE of the bathymetry reconstruction for these two test sets as a function of the distances for different input variables in \Cref{mahal}. The vertical axis in all figures is the RMSE of the bathymetry reconstruction using VEGAS. The horizontal axes are Mahalanobis distance between data points from either of these two test sets and the training set for easting velocity (left), northing velocity (middle), and the latent space (right). As the distance of test set data points become larger (orange cluster compared to blue cluster), the RMSE also increases, as expected. By comparing the difference between RMSE and Mahalanobis distance of the mean of the two clusters, we observe that doubling distance has approximately led to RMSEs twice as large. 

\begin{figure}[htbp]
\centering
\begin{subfigure}{1.0\textwidth}
\centering
\includegraphics[width=1.0\linewidth]{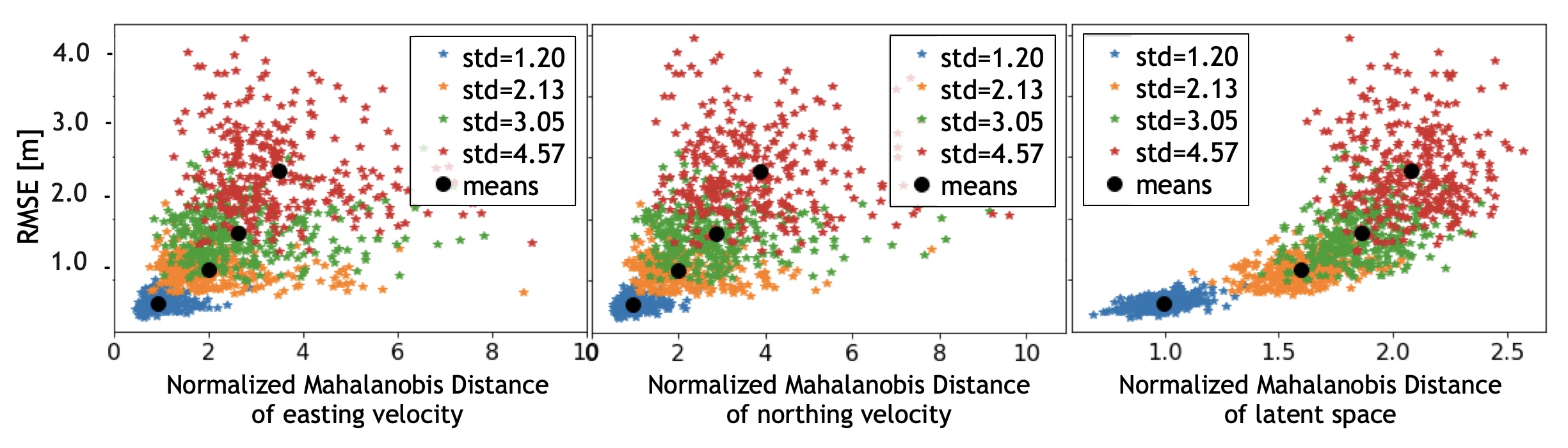}
%\caption{}
\label{ab_initiomodel}
\end{subfigure}
\caption{Comparison between the RMSE of the reconstructed bathymetries with the test set ``Parabolic +  Gaussian kernel'' with different standard deviations as a function of the Mahalanobis distances of easting velocity (left), northing velocity (middle), and latent space (c). The training set here has a standard deviation of 1.2 m (similar to the test dataset shown in blue), and the distances are calculated with respect to this distribution.% %The ratio between the means of the RMSEs and distances are approximately preserved for the clusters (left and middle subfigures).
}
\label{mahal_std}
\end{figure}

We have plotted similar RMSE-Mahalanobis distance comparisons for test sets with the same distribution as the training set with variable standard deviations in \Cref{mahal_std}. In particular, we have generated test sets with distributions similar to training set distribution, ``parabolic +  Gaussian kernel'' distribution, with different standard deviation values, and have plotted RMSE-Mahalanobis distance plots for them, similar to the plots of \Cref{mahal}. The standard deviation of the bathymetries in the training set is 1.2 m (the blue points), while the other test sets have the standard deviations of 2.13, 3.05, and 4.57 m. As expected, the larger the standard deviation is, the larger the RMSE in the bathymetry reconstruction. We also observe that in general, similar to the plots in \Cref{mahal}, the ratio between RMSEs and Mahalanobis distance of flow velocities is approximately maintained (mean points in the left and middle figures).

\section{Conclusion}
\label{conclusion}

In this work, we have presented a framework for fast estimation of riverine bathymetry from flow velocity measurements with variable boundary conditions, by combining a DNN as a ROM with a Bayesian approach for inversion. Once the neural network is trained after the offline stage, the predictions (inversion) can be done in less than a few minutes, making online bathymetry reconstruction possible. Our results demonstrate a computational efficiency about three orders of magnitude faster than common inversion algorithms such as the PCGA. Furthermore, our algorithm, VEGAS, has greater flexibility in terms of new velocity observations, due to both its nonlinear dimension reduction technique and its Bayesian viewpoint to the inversion problem.

The use of a DNN-based forward solver instead of numerical solvers like AdH has led to significant speed-ups in the inversion iteration process. Our forward solver (the SVE) is about three orders of magnitude faster than the AdH based on our previous observations~\cite{SWE_forward}. Furthermore, using SVE in our DNN architecture introduces a powerful tool that captures nonlinearities present in the forward dynamics. Comparisons of VEGAS with the PCA-based solver show its significantly higher accuracy thanks to its nonlinear dimension reduction capability.

Adding a Gaussian kernel to the PCGA posterior estimate of the bathymetry (Section \ref{data_prep}) leads to generating a training set distribution that is relevant to the river under study. However, as it was discussed and shown in Section \ref{unseen_distb}, even when no such information is available and the training set is generated from a simple parabolic mean bathymetry profile, the inversion can produce relatively accurate bathymetries when applied to data points sampled from different unknown distributions. This is very important for construction of VEGAS models where no flow velocity/bathymetry measurements are available.

We have also shown that the inversion algorithm is capable of reasonably accurate reconstruction of the bathymetries even with sparse flow velocity measurements. This is a very important feature of the algorithm for practical applications, since in many cases, no high-resolution measurement of the flow velocity of the river is available, and instead the algorithm has to use the available (sparse) measurements to estimate the bathymetry. We have observed (Section \ref{global_partial}) that even with 0.5\% of the measurements, the algorithm is capable of capturing important features of the bathymetry.

The bathymetries provided to our DNNs were obtained by augmenting the posterior distribution of the PCGA inversion (the Gaussian kernel and the scaling factor). In the future, we will study incorporation of more complex and general distributions of bathymetries into our training sets, thus generalizing the forward (and consequently, inverse) solver to a larger class of bathymetries. Another assumption in this work is that the geometry of the reach was fixed. Important extensions include updating the methodology to allow for significant changes in the lateral geometry (due for example to bank overflow) and extending the approach to allow application to multiple classes of rivers.

Finally, for the cases with sparse flow velocity measurements, here we have studied situations where the measurements are available at equally-distance grid points. However, depending on the uncertainty observed in the bathymetry reconstruction, one should be able to find specific cross sections or measurement locations that are more influential in reducing the error in predictions in order to guide the collection of new flow velocity data. This would fit naturally into design of experiments approaches~\cite{DoE} and is an area of future interest. % For such studies (and broadly inversion problem), inversion techniques other than MAP, such as Markov chain Monte Carlo (MCMC) algorithms, can be explored in future, due to significant speed-up of the forward solver thanks to the SVE.

\section*{Code sources}

The PCGA codes can be found in the \href{https://github.com/jonghyunharrylee/pyPCGA}{https://github.com/jonghyunharrylee/pyPCGA} github repository, and the VEGAS codes can be found in the \href{https://github.com/moji1369/VEGAS}{https://github.com/moji1369/VEGAS} github repository.

\section*{Acknowledgements}

This research was supported by the U.S. Department of Energy, Office of Advanced Scientific Computing Research under the Collaboratory on Mathematics and Physics-Informed Learning Machines for Multiscale and Multiphysics Problems (PhILMs) project, PhILMS grant DE-SC0019453. This work was also supported by an appointment to the Faculty and Postdoctoral Fellow Research Participation Program at the U.S. Engineer Research and Development Center, Coastal and Hydraulics Laboratory administered by the Oak Ridge Institute for Science and Education through an inter-agency agreement between the U.S. Department of Energy and ERDC. The Chief of Engineers has granted permission for the publication of this work.

\printbibliography
\end{document}